\documentclass[runningheads]{llncs}
\usepackage[T1]{fontenc}
\usepackage{graphicx,verbatim}
\usepackage{multirow}
\usepackage{booktabs}
\usepackage{diagbox}
\usepackage{bbm}
\usepackage{amsmath}
\usepackage{amssymb} 
\usepackage{tabularx}
\usepackage{array}
\usepackage{url}
\usepackage[misc]{ifsym}

\begin{document}
\title{VISTA: Variance-Gated Inter-Sequence Test-Time Adaptation for Multi-Sequence MRI Segmentation}
%

\author{
Zhipeng Deng\inst{1} \and
Jiale Zhou\inst{1} \and
Wenhan Jiang\inst{1} \and
Haolin Wang\inst{2} \and
Xun Lin\inst{3} \and
Yafei Ou\inst{4} \textsuperscript{\Letter} \and
Yefeng Zheng\inst{1}\textsuperscript{\Letter}
}







%
\authorrunning{Z. Deng et al.}
\institute{
Westlake University, Hangzhou, China
\and
Hokkaido University, Sapporo, Japan
\and
The Chinese University of Hong Kong, Hong Kong SAR, China
\and
RIKEN, Japan
\\
\email{
\{dengzhipeng, zhoujiale, jiangwenhan, zhengyefeng\}@westlake.edu.cn;
yafei.ou@riken.jp
}
}

\titlerunning{VISTA} 

\maketitle              




%

\begin{abstract}
Deploying multi-sequence magnetic resonance imaging (MRI) segmentation models to new clinical environments is challenging due to variations in scanners and acquisition protocols. 
Although existing TTA methods handle basic per-modality shifts, they often fail under a fundamental dual-shift problem, as their adaptation signals fail to capture modality-interaction shifts that disrupt inter-sequence consistency.
To address this, we propose \textbf{V}ariance-gated \textbf{I}nter-\textbf{S}equence \textbf{T}est-time \textbf{A}daptation (VISTA), a source-free framework that tackles modality-interaction shifts.
First, we design an Inter-Sequence Intervention Generator (ISIG) that generates a set of consistency probes by swapping low-frequency spectra and entropy-localized patches across sequences, preserving anatomical semantics while challenging inter-sequence dependencies.
Second, we introduce Cross-View Disagreement-Aware Pseudo Labeling (CDPL), which establishes a voxel-wise reliability metric using cross-view disagreement variance to dynamically gate self-training and enforce interventional consistency, encouraging the network to rely on robust anatomical semantics.
Extensive experiments adapting from standard adult MRI (BraTS-GLI-Pre) to African low-field (BraTS-SSA) and pediatric (BraTS-PED) cohorts show improved performance over competing methods under clinical shifts, achieving absolute Dice improvements of +1.89\% (SSA) and +2.82\% (PED) over the source model. The code is available at \url{https://github.com/dzp2095/VISTA}.

\end{abstract}

\section{Introduction}
\begin{figure}
    \centering
    \includegraphics[width=\linewidth]{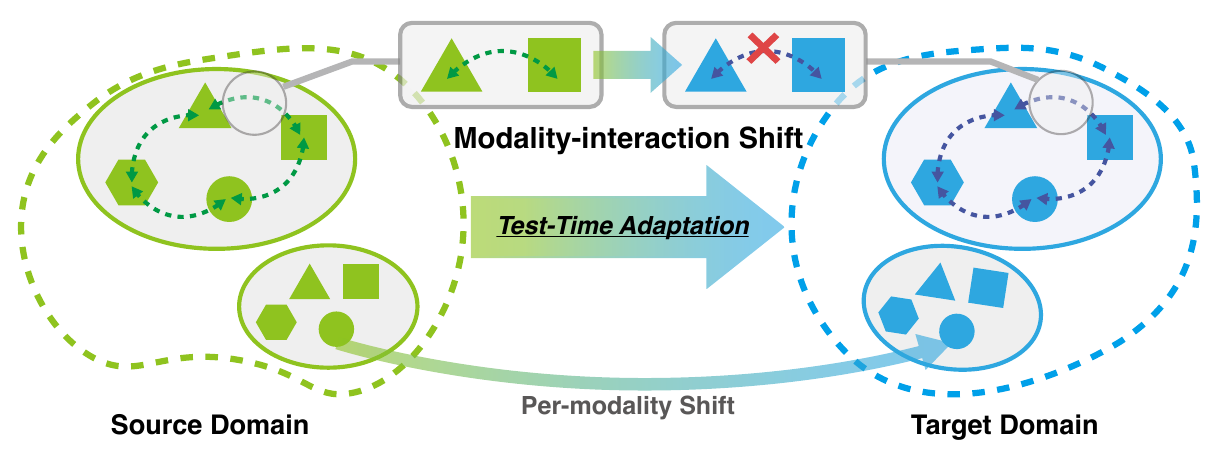}
    \caption{\textbf{The dual-shift problem in multi-sequence MRI.} Beyond conventional \emph{per-modality shift} (degradation within individual sequences), clinical deployment fundamentally suffers from \emph{modality-interaction shift}—the disruption of inherent cross-sequence structural dependencies.}
    \label{fig:motivation}
\end{figure}

Multi-sequence MRI segmentation is a cornerstone for clinical diagnosis, as different sequences (e.g., T1, T2, FLAIR) provide complementary tissue evidence~\cite{menze2015brats,bakas2017tcga_radiomics}. When models are deployed to new sites or scanners, they often suffer from severe performance degradation~\cite{zech2018generalization,deng2026fedsemidg,you2026towards}, motivating \emph{test-time adaptation} (TTA) using unlabeled target data. However, the fundamental problem limiting the success of current TTA in multi-modal settings is an oversimplified view of ``domain shift''. In multi-sequence MRI, domain shift is not a simple distribution change, but a critical \textbf{dual-shift problem} (Fig.~\ref{fig:motivation}): \emph{per-modality} and \emph{modality-interaction} shifts. While per-modality shift alters isolated sequence statistics, modality-interaction shift critically disrupts the structural mapping where complementary sequences corroborate anatomical boundaries. When this consistency breaks down, traditional adaptation signals become unreliable.

The success of test-time adaptation fundamentally depends on mining reliable supervision signals from unlabeled target data. To extract these signals, existing paradigms optimize generic confidence via entropy minimization~\cite{wang2021tent,niu2022eata,zhang2025come}, test-time normalization~\cite{karani2021ttan,kang2024membn}, or prompting~\cite{zhang2024testfit,chen2024vptta,zhang2025pass}, alongside exploring robust segmentation and consistency~\cite{wang2025smart,joshi2025muvi,yang2020fda,zakazov2022feather,zhou2025topotta}. Designed primarily for single-modal tasks, they treat multi-sequence inputs as a unified whole, overlooking asymmetric sequence degradation. While recent multi-modal TTA methods address modality-level reliability to prevent negative transfer~\cite{cao2023mm_continual_tta,chen2025selective_unimodal_shift}, they ignore the \emph{spatially localized, voxel-wise} contradictions inherent to co-registered MRI that demand fine-grained reliability control.
Consequently, when modality-interaction shifts disrupt cross-contrast relationships, optimizing adaptation objectives over the resulting contradictory regions forces the network to fit flawed evidence, driving rapid error accumulation and adaptation collapse \cite{wang2022cotta,yu2025medsegtta}.

To address this critical gap, we propose \textbf{VISTA} (\textbf{V}ariance-gated \textbf{I}nter-\textbf{S}equence \textbf{T}est-time \textbf{A}daptation), which mitigates modality-interaction shifts by transforming inter-sequence corroboration into a reliable adaptation signal. Specifically, an Inter-Sequence Intervention Generator (ISIG) actively probes structural dependencies via controlled cross-sequence perturbations, enabling our Cross-View Disagreement-Aware Pseudo Labeling (CDPL) to dynamically gate voxel-wise self-training using the resulting variance. Extensive experiments adapting from standard adult MRI (BraTS-GLI-Pre) to African low-field (BraTS-SSA) and pediatric (BraTS-PED) cohorts confirm that VISTA effectively mitigates severe degradation and shows improved performance over competing methods, achieving absolute Dice improvements of +1.89\% and +2.82\%, respectively.

\begin{figure}[t!]
    \centering
    \includegraphics[width=\linewidth]{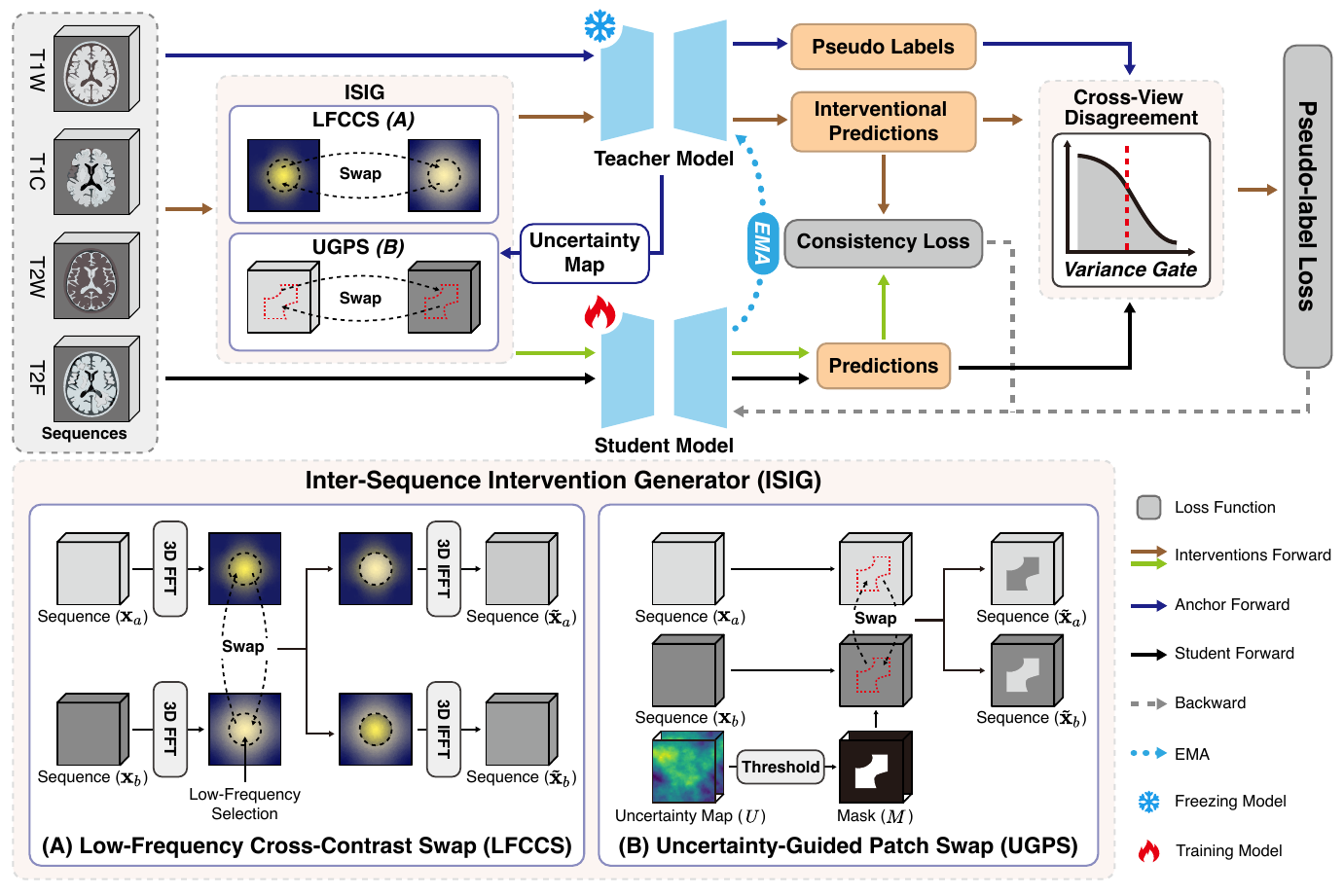}
    \caption{\textbf{The pipeline of our VISTA.}  ISIG simulates modality-interaction shifts via frequency (LFCCS) and spatial (UGPS) cross-sequence swaps. The resulting cross-view disagreement variance dynamically gates the teacher's anchor pseudo-labels, preventing error accumulation and providing reliable supervision for student optimization.}
    \label{fig:framework}
\end{figure}

\section{Method}

\subsection{Problem Setup}
We study source-free test-time adaptation (TTA) for multi-sequence MRI segmentation. Let $\mathbf{x} \in \mathbb{R}^{S \times H \times W \times D}$ denote a co-registered $S$-sequence MRI volume, and $f_{\theta_0}$ be a segmentation network pre-trained on a labeled source domain. For clarity, we denote the pre-sigmoid logits as $\mathbf{z} = f_{\theta}(\mathbf{x})$ and probabilities as $\mathbf{p} = \sigma(\mathbf{z})$. At deployment, given a continuous stream of unlabeled target volumes $\{\mathbf{x}_t\}_{t=1}^{T}$ , TTA aims to minimize the unknown target risk online. We achieve this by constructing an unsupervised objective $\mathcal{L}(\theta;\mathbf{x}_t)$ on-the-fly to perform gradient updates at test time~\cite{wang2021tent,niu2022eata}.

\subsection{Inter-Sequence Intervention Generator (ISIG)}
ISIG generates a small set of \emph{inter-sequence intervention views} that act as consistency probes. These interventions are designed to preserve anatomical semantics while actively challenging the network's inter-sequence dependencies. 
Let $\mathbf{x}^{(0)} = \mathbf{x}$ be the original input volume, comprising $S$ sequences denoted as $\{\mathbf{x}_s\}_{s=1}^S$. ISIG produces $K_v$ views, where each view retains all $S$ sequences but introduces targeted cross-sequence perturbations.

\noindent\textbf{Low-Frequency Cross-Contrast Swap (LFCCS).} Motivated by findings that low-frequency Fourier amplitudes dictate appearance while phase preserves structure~\cite{yang2020fda,zakazov2022feather}, we introduce a sequence-level intervention tailored for MRI. Since multi-sequence MRI shares identical anatomy but differs in contrast, LFCCS generates perturbations by mutually exchanging the low-frequency spectra \emph{between sequences} rather than across different subjects. This mechanism effectively simulates severe intensity shifts and challenges modality interactions while strictly preserving anatomical boundaries. 

Let $\mathcal{F}$ denote the 3D Fast Fourier Transform. Recall that $\mathbf{x}^{(0)}$ is the original unperturbed input. For any sequence $\mathbf{x}_s^{(0)} \in \{\mathbf{x}_s^{(0)}\}_{s=1}^S$, its amplitude and phase spectra are explicitly defined as $A_s^{(0)} = |\mathcal{F}(\mathbf{x}_s^{(0)})|$ and $\Phi_s^{(0)} = \angle \mathcal{F}(\mathbf{x}_s^{(0)})$. Let $M_{r}$ be a centered low-frequency binary mask with a bandwidth ratio $r \in (0,1)$. During each intervention step, we randomly select an arbitrary sequence pair $(a, b)$ to perform a mutual swap of their original amplitudes:
\begin{equation}
\begin{aligned}
\tilde{A}_a &= (1-M_{r})\odot A_a^{(0)} + M_{r}\odot A_b^{(0)}, &\quad \tilde{\mathbf{x}}_a &= \mathcal{F}^{-1}\!\left(\tilde{A}_a \cdot e^{\mathrm{i}\Phi_a^{(0)}}\right), \\
\tilde{A}_b &= (1-M_{r})\odot A_b^{(0)} + M_{r}\odot A_a^{(0)}, &\quad \tilde{\mathbf{x}}_b &= \mathcal{F}^{-1}\!\left(\tilde{A}_b \cdot e^{\mathrm{i}\Phi_b^{(0)}}\right),
\end{aligned}
\end{equation}
while keeping the remaining sequences unchanged. This perturbed multi-sequence volume yields our first interventional view, denoted as $\mathbf{x}^{(1)}$.

\noindent\textbf{Uncertainty-Guided Patch Swap (UGPS).}
While LFCCS provides a global contrast perturbation, modality-interaction shifts also occur locally. We treat high predictive entropy as a cue for spatial locations where the model may struggle to fuse conflicting multi-sequence evidence. To target these exact vulnerabilities, we extend the concept of label-preserving patch mixing~\cite{yun2019cutmix} into an active multi-sequence probe. Specifically, UGPS restricts cross-sequence patch swaps exclusively to teacher-identified high-entropy regions, dynamically exchanging local evidence where the model is most prone to collapse.

To localize these regions, we first obtain an anchor prediction from the teacher, $\mathbf{p}^{(0)} \in [0,1]^{C \times H \times W \times D}$, and compute the voxel-wise entropy map $U$ via channel-averaged binary predictive entropy. We then derive a binary mask $M$ by thresholding $U$ at its $q$-th quantile ($\tau_q$) and applying a 3D morphological dilation to cover the surrounding structural context:$M = \mathrm{Dilate}\big(\mathbbm{1}[U \ge \tau_q]\big)$.

Given a randomly selected sequence pair $(a, b)$, UGPS spatially mixes their original unperturbed sequence views ($\mathbf{x}_a^{(0)}$ and $\mathbf{x}_b^{(0)}$) to our second probe view, $\mathbf{x}^{(2)}$ based on the  high-entropy mask $M$:
\begin{equation}
\begin{aligned}
\tilde{\mathbf{x}}_a &= (1-M)\odot \mathbf{x}_a^{(0)} + M\odot \mathbf{x}_b^{(0)}, \\
\tilde{\mathbf{x}}_b &= (1-M)\odot \mathbf{x}_b^{(0)} + M\odot \mathbf{x}_a^{(0)}.
\end{aligned}
\end{equation}

\noindent\textbf{View Set Construction.} 
Consequently, ISIG outputs a view set: $\{\mathbf{x}^{(k)}\}_{k=0}^{2} = \{\mathbf{x}^{(0)}, \mathbf{x}^{(1)}, \mathbf{x}^{(2)}\}$, representing the original input, the LFCCS-perturbed view, and the UGPS-perturbed view, respectively. For the sequence pair selection $(a,b)$, we randomly sample a pair of distinct sequences for both LFCCS and UGPS for each forward pass to ensure diverse intervention dynamics.

\subsection{Cross-View Disagreement-Aware Pseudo Labeling (CDPL)}
CDPL converts the prediction stability across ISIG views into a voxel-wise reliability signal for self-training.

\noindent\textbf{Cross-View Disagreement Variance.}
We first compute the teacher's voxel-wise probability predictions $\mathbf{p}^{(k)}$ for each view $k \in \{0, 1, 2\}$. To quantify the model's sensitivity to modality-interaction shifts, we compute the \emph{disagreement variance} $V$ across these predictions:
\begin{equation}
V_c(\mathbf{v}) = \mathrm{Var}_{k}\big(p_{c}^{(k)}(\mathbf{v})\big),
\end{equation}
where $\mathrm{Var}_{k}(\cdot)$ denotes the variance calculated across the three generated views, $\mathbf{v}$ denotes the 3D spatial voxel coordinate, and $c$ is the class channel. Intuitively, a large $V_c(\mathbf{v})$ indicates that the prediction at this voxel is highly sensitive to cross-sequence perturbations, making it unreliable for self-training. Using prediction variance across intervention views provides a perturbation-stability measure for pseudo-label reliability, following the intuition that predictions stable under anatomy-preserving perturbations are more reliable for self-training.


\noindent\textbf{Variance-Gated Pseudo-Labeling.}
We derive self-training supervision exclusively from the unperturbed teacher anchor probabilities $\mathbf{p}^{(0)}$. To optimize the model, we extract the corresponding logits $\mathbf{z}_{\theta}^{(0)}$ from the active student network $f_{\theta}$. Rather than only trusting confident predictions, we employ a strict variance gate to filter out structurally contradictory evidence. The variance-gated pseudo-label loss is formulated as a joint-masked objective:
\begin{equation}
\mathcal{L}_{\mathrm{PL}} = \sum_{\mathbf{v},c} \underbrace{\mathbbm{1}\big[V_{c}(\mathbf{v}) \le \tau_{\mathrm{var}}\big]}_{\text{Variance Gate}} \cdot \underbrace{\mathbbm{1}\big[p_{c}^{(0)}(\mathbf{v}) \notin (\tau^{-}, \tau^{+})\big]}_{\text{Confidence Gate}} \cdot \ell\big(z_{\theta, c}^{(0)}(\mathbf{v}), \tilde{y}_{c}(\mathbf{v})\big),
\end{equation}
where $\tilde{y}_{c}(\mathbf{v}) \in \{0, 1\}$ is the hard pseudo-label (set to $1$ if $p_{c}^{(0)}(\mathbf{v}) \ge \tau^{+}$ and $0$ if $p_{c}^{(0)}(\mathbf{v}) \le \tau^{-}$), and $\ell(\cdot,\cdot)$ is the standard binary cross-entropy loss. 

\noindent\textbf{Consistency Regularization.}
Fundamentally, multi-sequence models are vulnerable to clinical shifts because they overfit to specific inter-sequence correlations. Since our ISIG module explicitly simulates these modality-interaction shifts while preserving the underlying anatomy, we leverage it to enforce a strict invariance constraint. By aligning the student's predictions on the intervention views with the unperturbed teacher anchor:
\begin{equation}
\mathcal{L}_{\mathrm{CONS}} = \frac{1}{2} \sum_{k=1}^{2} d\big(\sigma(\mathbf{z}_{\theta}^{(k)}), \mathrm{stopgrad}(\mathbf{p}^{(0)})\big),
\end{equation}
where $d(\cdot, \cdot)$ computes the soft binary cross-entropy (BCE) loss. This objective compels the network to base its predictions on robust anatomical semantics rather than fragile modality interactions.

\subsection{Overview of the VISTA Framework}
Integrating the variance-gated pseudo-labeling and the explicit invariance constraint, the overall unsupervised adaptation objective for VISTA is defined as $\mathcal{L} = \mathcal{L}_{\mathrm{PL}} + \lambda \mathcal{L}_{\mathrm{CONS}}$, where $\lambda$ is a hyperparameter balancing the two terms. 

During deployment, VISTA operates under an online continual setting. For each incoming unlabeled multi-sequence volume from the target stream, we perform gradient updates on the student network $f_{\theta}$ to minimize $\mathcal{L}$. Simultaneously, the teacher network $f_{\bar{\theta}}$ is smoothly updated via an exponential moving average (EMA) of the student's weights, and its unperturbed output is utilized to generate the final stable segmentation mask.

\section{Experiments}

\subsection{Datasets and Experiment Setup}

\noindent\textbf{Datasets and Shift Setting.} 
We train the source model on the standard BraTS-GLI-Pre (2025 edition, 1251 cases) cohort~\cite{menze2015brats,bakas2017tcga_radiomics}, of standard adult pre-operative gliomas. To evaluate adaptation robustness under severe clinical shifts, we deploy the model on two challenging target cohorts: (1) the BraTS-SSA cohort (2023 edition, 60 public training cases) \cite{adewole2025brats}, which captures severe acquisition heterogeneity and resource-limited low-field imaging conditions; and (2) the BraTS-PED cohort (2024 edition, 261 public training cases) \cite{kazerooni2024bratspeds}, introducing distinct pediatric tumor characteristics and multi-institutional variability. 

\noindent\textbf{Preprocessing.} All volumes are provided already co-registered. We perform skull-stripping for the BraTS-PED cohort using a standard brain extraction tool (i.e., HD-BET) to ensure anatomical consistency. All volumes are resized to $160 \times 192 \times 160$ voxels and independently standardized using voxel-wise Z-score normalization. To evaluate the model consistently across domains, we harmonize the original annotations into the standard BraTS nested subregions: Enhancing Tumor (ET), Tumor Core (TC), and Whole Tumor (WT). 

\noindent\textbf{Evaluation Metrics.} 
Following the standard BraTS evaluation protocol, we compute the Dice Similarity Coefficient (Dice), 95\% Hausdorff Distance (HD95), and Sensitivity for the ET, TC, and WT subregions, reporting their macro-averages.  All experiments are repeated three times. To save space, Table~\ref{tab:main} reports mean values, while Table~\ref{tab:ablation} reports mean$\pm$standard deviation.

\noindent\textbf{Implementation Details.} 
We employ the standard 3D U-Net \cite{cicek2016_3dunet} backbone implemented in MONAI~\cite{cardoso2022monai}. During deployment, VISTA operates under a strict online continual adaptation protocol without revisiting the source data. SSA and PED are evaluated as separate continual deployment streams, each initialized from the same source model and adapted sequentially on incoming unlabeled target volumes.  For each incoming target volume (batch size 1), we perform $K=10$ gradient update steps on the student network using the Adam optimizer with a learning rate of $10^{-4}$. Following standard parameter-efficient TTA practices, we restrict the gradient updates exclusively to the affine parameters ($\gamma$ and $\beta$) of the Batch Normalization layers. The teacher network is smoothly updated via EMA with a decay rate of $\alpha=0.99$. 
For the ISIG module, we use $K_v=3$ views, setting the low-frequency mask ratio to $r=0.10$ for LFCCS, and thresholding the entropy map at the $q=0.95$ quantile with a $15 \times 15 \times 15$ dilation kernel for UGPS. For the CDPL module, the variance gate threshold is set to $\tau_{\mathrm{var}}=0.05$, and the pseudo-label confidence thresholds are $\tau^{+}=0.95$ and $\tau^{-}=0.05$. The consistency regularization weight is set to $\lambda=1.0$. 
To strictly adhere to the source-free setting, all hyperparameters are selected once on a 20$\%$ held-out source-domain validation set before target deployment and then kept fixed for all target cohorts, without using any target-domain annotations or target-specific validation feedback.  All experiments are conducted on a single NVIDIA RTX 5090 GPU.

\subsection{Results and Analysis}
As shown in Table~\ref{tab:main}, VISTA demonstrates superior and consistent adaptation performance across both clinical targets. A critical challenge in test-time adaptation is the risk of negative transfer, where adapting on shifted or noisy target data degrades performance below the source pre-trained model (No TTA). This vulnerability is clearly evident in our experiments: on the SSA cohort, several baselines such as Tent, EATA, and SmaRT fail to surpass the No TTA baseline (78.42\% Dice). Similarly, methods like VPTTA, PASS, and TestFit suffer from degradation on the PED cohort. In contrast, VISTA effectively prevents such collapse, achieving the highest Dice scores on both the SSA (80.31\%) and PED (42.03\%) datasets. This stability confirms that our cross-sequence disagreement variance gate successfully filters out structurally contradictory pseudo-labels, ensuring safe and robust continual learning.

The advantage of VISTA is particularly pronounced under the extreme clinical shifts presented by the pediatric (PED) cohort. While conventional methods struggle with the variability of pediatric tumors, VISTA outperforms strong baselines like Tent and MemBN. This improvement indicates that by explicitly penalizing sensitivity to intervention views ($\mathcal{L}_{\mathrm{CONS}}$), our framework successfully forces the network to rely on robust anatomical semantics rather than fragile modality correlations. Furthermore, VISTA excels in spatial precision and outlier suppression. It achieves the lowest 95\% Hausdorff Distance (HD95) across both cohorts (20.15 mm on SSA and 50.88 mm on PED) alongside the highest Sensitivity. This demonstrates that rather than producing fragmented masks, our method preserves sharp, anatomically plausible tumor boundaries even under severe clinical shifts.

\begin{table}[t!]
\centering
\caption{Macro-average over ET/TC/WT on SSA and PED targets. Higher is better for Dice and Sensitivity (\%), lower is better for HD95 (mm). The best results are \textbf{bolded} and the second-best are \underline{underlined}.}
\label{tab:main}
\begin{tabular}{lcccccc}
\toprule
& \multicolumn{3}{c}{SSA Target} & \multicolumn{3}{c}{PED Target} \\
\cmidrule(lr){2-4}\cmidrule(lr){5-7}
Method & Dice $\uparrow$ & Sens. $\uparrow$ & HD95 $\downarrow$ & Dice $\uparrow$ & Sens. $\uparrow$ & HD95 $\downarrow$ \\
\midrule
No TTA & 78.42 & 78.41 & 21.34 & 39.21 & 55.91 & 54.77 \\
\midrule
Tent~\cite{wang2021tent} & 77.82 & 78.07 & 21.42 & \underline{40.96} & 55.32 & 53.06 \\
EATA~\cite{niu2022eata} & 76.96 & 78.63 & \underline{21.09} & 39.25 & 55.48 & 52.35 \\
COME~\cite{zhang2025come} & 77.84 & 79.30 & 21.86 & 40.17 & 56.41 & \underline{51.42} \\
VPTTA~\cite{chen2024vptta} & \underline{79.85} & 81.04 & 21.70 & 38.31 & \underline{56.87} & 55.99 \\
PASS~\cite{zhang2025pass} & 79.73 & \underline{81.37} & 22.45 & 38.68 & 56.69 & 52.64 \\
TestFit~\cite{zhang2024testfit} & 77.79 & 77.49 & 21.91 & 36.36 & 44.48 & 53.15 \\
MemBN~\cite{kang2024membn} & 78.75 & 79.72 & 21.26 & 40.86 & 56.15 & 52.35 \\
SmaRT~\cite{wang2025smart} & 72.17 & 67.75 & 24.64 & 37.25 & 48.78 & 53.92 \\
\midrule
\textbf{VISTA (Ours)} & \textbf{80.31} & \textbf{82.87} & \textbf{20.15} & \textbf{42.03} & \textbf{57.39} & \textbf{50.88} \\
\bottomrule
\end{tabular}
\end{table}

\subsection{Ablation Studies}
To evaluate the contribution of each proposed component, we conduct a detailed ablation study on the challenging PED cohort, as reported in Table~\ref{tab:ablation}. Compared to the No TTA baseline (39.21\% Dice), applying either standard pseudo-labeling (PL) or consistency regularization (CONS) individually yields only marginal improvements. While directly combining them without gating (PL + CONS) improves the Dice score to 41.43\%, it remains sub-optimal due to the unrestricted confirmation bias caused by severe modality-interaction shifts. Crucially, introducing our cross-sequence disagreement variance gate (Ours Full) boosts the performance to the peak of 42.03\%, demonstrating that strictly filtering structurally contradictory evidence is essential for safe self-training. Furthermore, removing either the UGPS or LFCCS interventional view degrades the performance to 40.92\% and 41.47\% respectively, confirming that both frequency-level style swaps and spatial-level entropy-guided patch swaps are indispensable for comprehensively simulating clinical shifts.

\begin{table}[t!]
\centering
\caption{Ablation study of the proposed framework. The ISIG module comprises two intervention views: ``LFCCS'' and ``UGPS''. ``CONS'' denotes consistency regularization, ``PL'' stands for pseudo-labeling, and ``Var Gate'' represents the cross-sequence disagreement variance gating. The best result is \textbf{bolded}.}
\label{tab:ablation}
\begin{tabular}{l|cc|ccc|c}
\hline
\multirow{2}{*}{Variant} & \multicolumn{2}{c|}{ISIG} & \multirow{2}{*}{CONS} & \multirow{2}{*}{PL} & \multirow{2}{*}{Var Gate} & \multirow{2}{*}{Dice $(\%)\uparrow$} \\
\cline{2-3}
 & LFCCS & UGPS & & & & \\
\hline
No TTA    &     &       &      &     &     & 39.21$\pm$1.78 \\
PL only    &     &   &     & \checkmark &    & 39.76$\pm$1.09 \\
CONS only    & \checkmark & \checkmark & \checkmark &    &    & 40.19$\pm$0.78 \\
PL + CONS (no gate)    & \checkmark & \checkmark & \checkmark & \checkmark &       & 41.43$\pm$0.94 \\
\hline
Ours (w/o UGPS)   & \checkmark &  & \checkmark & \checkmark & \checkmark & 40.92$\pm$1.29 \\
Ours (w/o LFCCS)   &   & \checkmark & \checkmark & \checkmark & \checkmark & 41.47$\pm$0.73 \\
\textbf{Ours (Full)}   & \checkmark & \checkmark & \checkmark & \checkmark & \checkmark & \textbf{42.03}$\pm$1.14 \\
\hline
\end{tabular}
\end{table}

\section{Conclusion}
We presented a simple and reliable source-free test-time adaptation framework for multi-sequence MRI segmentation.
By generating inter-sequence intervention views and using cross-view disagreement variance to gate pseudo-label supervision, VISTA improves robustness under severe cohort shifts such as SSA and pediatric targets.
One limitation is that VISTA assumes approximately aligned sequences and uses fixed thresholds selected on source validation; abrupt switching between target domains within a single continual stream also remains unexplored.
Future work will explore alignment-aware interventions, adaptive thresholding, domain-shift detection or reset strategies, and broader multi-sequence clinical applications such as prostate and cardiac MRI.

\begin{credits}

\subsubsection{\ackname}
This work was supported by Zhejiang Leading Innovative and Entrepreneur Team Introduction Program (2024R01007)  and the “Pioneer” and “Leading Goose” Research and Development Program of Zhejiang (2025C02077). 

\subsubsection{\discintname}
The authors have no competing interests to declare that are relevant to the content of this article.
\end{credits}

%
%
%
%

\bibliographystyle{splncs04}
\bibliography{reference}

\end{document}